\documentclass[preprint]{vgtc}               % preprint
\ifpdf%                                % if we use pdflatex
  \pdfoutput=1\relax                   % create PDFs from pdfLaTeX
  \usepackage{graphicx}                % allow us to embed graphics files
  \DeclareGraphicsExtensions{.pdf,.png,.jpg,.jpeg} % for pdflatex we expect .pdf, .png, or .jpg files
\else%                                 % else we use pure latex
  \usepackage{graphicx}                % allow us to embed graphics files
  \DeclareGraphicsExtensions{.eps}     % for pure latex we expect eps files
\fi%

\graphicspath{{res/images/}} % where to search for the images

\usepackage{microtype}                 % use micro-typography (slightly more compact, better to read)
\PassOptionsToPackage{warn}{textcomp}  % to address font issues with \textrightarrow
\usepackage{textcomp}                  % use better special symbols
\usepackage{mathptmx}                  % use matching math font
\usepackage{times}                     % we use Times as the main font
\usepackage{cite}                      % needed to automatically sort the references
\usepackage{tabu}                      % only used for the table example
\usepackage{booktabs}                  % only used for the table example
\usepackage{enumitem}
\usepackage{amsmath}
\usepackage{amssymb}
\usepackage{bbm}
\usepackage{epsfig}
\usepackage{makecell}
\usepackage{multirow}
\usepackage{xcolor}
\usepackage{color}
\usepackage{amsfonts}
\def\ie{i.e., }
\def\etal{et al.}

\onlineid{3885}

\vgtccategory{Technology}

\vgtcinsertpkg
\usepackage{subcaption}

\preprinttext{To appear in IEEE International Symposium on Mixed and Augmented Reality (ISMAR), Singapore, Oct 2022.}

\title{Temporal View Synthesis of Dynamic Scenes through 3D Object Motion Estimation with Multi-Plane Images}

\author{Nagabhushan Somraj\thanks{e-mail: nagabhushans@iisc.ac.in}%
\and Pranali Sancheti %
\and Rajiv Soundararajan\thanks{e-mail: rajivs@iisc.ac.in \newline Code and dataset available at \newline https://nagabhushansn95.github.io/publications/2022/DeCOMPnet.html}}
\affiliation{\scriptsize Department of Electrical Communication Engineering,\\Indian Institute of Science}

\teaser{
  \centering
  \includegraphics[width=\linewidth]{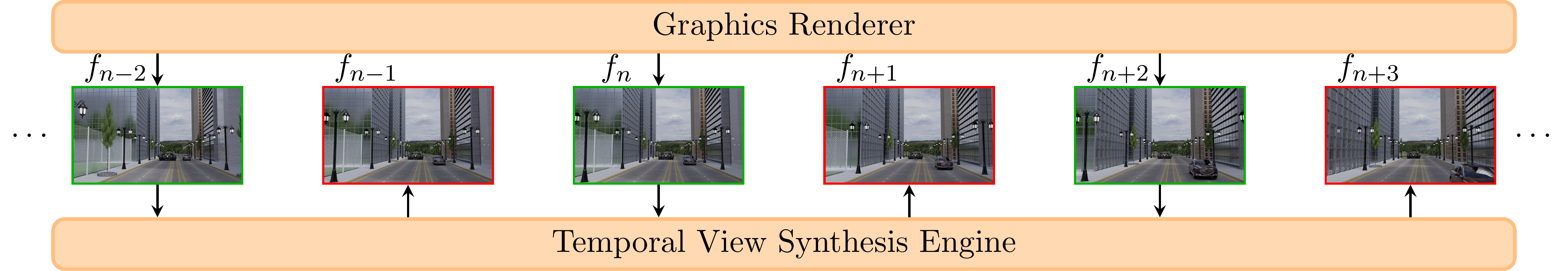}
  \caption{\textbf{Frame-rate upsampling of graphically rendered dynamic videos using Temporal View Synthesis.}
  This illustration shows upsampling by a factor of two.
  The graphics renderer renders alternate frames $ \left\{ f_{n-2}, f_n, f_{n+2}, \ldots \right\} $ and the intermediate frames $ \left\{ f_{n-1}, f_{n+1}, f_{n+3}, \ldots \right\} $ are predicted using temporal view synthesis.
  For better visualization of motion, we show frames which are 10 time instants apart instead of consecutive frames.}
  \label{fig:frame-rate-upsampling}
}

\abstract{
    The challenge of graphically rendering high frame-rate videos on low compute devices can be addressed through periodic prediction of future frames to enhance the user experience in virtual reality applications.
    This is studied through the problem of temporal view synthesis (TVS), where the goal is to predict the next frames of a video given the previous frames and the head poses of the previous and the next frames.
    In this work, we consider the TVS of dynamic scenes in which both the user and objects are moving.
    We design a framework that decouples the motion into user and object motion to effectively use the available user motion while predicting the next frames.
    We predict the motion of objects by isolating and estimating the 3D object motion in the past frames and then extrapolating it.
    We employ multi-plane images (MPI) as a 3D representation of the scenes and model the object motion as the 3D displacement between the corresponding points in the MPI representation.
    In order to handle the sparsity in MPIs while estimating the motion, we incorporate partial convolutions and masked correlation layers to estimate corresponding points.
    The predicted object motion is then integrated with the given user or camera motion to generate the next frame.
    Using a disocclusion infilling module, we synthesize the regions uncovered due to the camera and object motion.
    We develop a new synthetic dataset for TVS of dynamic scenes consisting of 800 videos at full HD resolution.
    We show through experiments on our dataset and the MPI Sintel dataset that our model outperforms all the competing methods in the literature.%
} % end of abstract

\CCScatlist{
  \CCScatTwelve{View synthesis}{Video prediction}{Interleaved reprojection}{3D optical flow}
}

\begin{document}


\firstsection{Introduction}

\maketitle

%% \section{Introduction} %for journal use above \firstsection{..} instead
    The computational limitations of handheld mobile devices reduce the frame rates at which high resolution video content can be graphically rendered in virtual reality (VR) applications.
    This leads to a poor user experience when viewing such content on a VR headset~\cite{oculus2015atw_examined}.
    To overcome this limitation, interleaved reprojection techniques~\cite{oculus2016atw, oculus2016asw, oculus2019asw2, valve2016interleaved} graphically render frames periodically and predict the intermediate frames.
    The prediction of these frames needs to adjust for the camera motion or change in view point of the user as well as the movement of objects in the scene.
    This leads us to the problem of temporal view synthesis of dynamic scenes (TVS-DS), where the goal is to generate the future video frame given its camera pose.
    
    Our problem statement is primarily motivated by use cases in gaming applications of VR.
    This is a significant share of VR applications. 
    TVS-DS is also applicable in other scenarios requiring real-time rendering such as flight simulations, VR exposure therapy and so on.
    %Further, TVS-DS can be employed to compress videos for streaming applications in VR.
    % Further, since TVS-DS predicts future frames, it can also be employed in video compression for streaming applications in VR\@.
    The above applications primarily involve graphical rendering of simulated environments and thus deal with synthetic videos. 
    In most of these situations, the content creators or game developers make the depth of the scene available along with other scene information required for rendering. 
    However, more detailed information such as motion vectors for moving objects may not typically be available.
    Thus, we particularly focus on predicting the future frames of a synthetic video given the past RGB-D frames and the head (or camera) pose for both the past and future frames.

    The main difference between TVS-DS and other related problems such as novel view synthesis~\cite{zhou2016view, wiles2020synsin} and depth image based rendering (DIBR)~\cite{cho2017hole,luo2020disocclusion} is the object motion prediction between source and target views.
    In the works by Gao \etal~\cite{gao2021dynamic}, Lin \etal~\cite{lin2021deep} and Yoon \etal~\cite{yoon2020novel}, the authors consider dynamic scene videos, but the target frame is at the same time instant as one of the source frames.
    Thus, they do not address the question of moving objects at a future time instant.
    Recently, HyperNeRF~\cite{park2021hypernerf} and the model by Li \etal~\cite{li2021neural} can interpolate object motion between the frames, but require hours of training for every scene.
    On the other hand, our problem formulation is close to that of video prediction~\cite{srivastava2015unsupervised,mathieu2016deep}, but differs in the use and availability of camera motion.
    The explicit use of camera motion can help TVS-DS methods perform much better than generic video prediction.
    Thus, TVS-DS lies at the intersection of video prediction and view synthesis.

    The key challenges in TVS-DS involve leveraging the user motion to extrapolate the past motion of moving objects, combining these to predict the next frame, and infilling any disocclusions arising out of the combined motion.
    In this work, we focus on causal frame-rate upsampling of graphically rendered videos as shown in \autoref{fig:frame-rate-upsampling} and explore the feasibility of using neural networks to predict future frames.
    Since it is vital that the infilling is temporally consistent with the next rendered frame, we do not aim to hallucinate new objects.
    While camera motion can lead to new regions entering the frame, this can be addressed by rendering a larger field of view and cropping the desired field of view while displaying.

%    \textcolor{red}{We build on the disentangled propagation and generation approach of Gao \etal~\cite{gao2019disentangling} and introduce two novel modifications for TVS-DS\@.
%    Firstly, we decouple the camera motion and local object motion in the past frames and only extrapolate the local motion to predict the movement of objects.
%    We train a deep optical flow network that operates on past frames compensated for camera motion to predict the future local flow of objects.
%    We also make effective use of the scene depth to compute the optical flow from the camera motion compensated frames.}
    We design a framework to decouple camera and object motion, which allows us to effectively use the available camera motion and only predict the object motion.
    To predict future object motion, we estimate the object motion in the past frames and then extrapolate it.
    However both the camera motion and object motion are intertwined in the past frames.
    To estimate the object motion alone, we first nullify the camera motion between the past frames by warping them to the same view using projective geometry.
    Decoupling camera and object motion makes the predicted object motion independent of the past or future camera motion, and thus we can synthesize future frames even when there is a change in the camera trajectory.

    The depth of moving objects in a scene is usually different from that of their neighboring pixels.
    This difference can be exploited to better estimate the object motion by matching the points in 3D instead of 2D\@.
    Driven by this observation, we propose a method to estimate object motion in 3D, which we show to be more accurate than 2D motion estimation.
%    We hypothesize that estimating object motion in 3D provides a more accurate motion estimate than in 2D, since different objects are typically farther apart in 3D leading to easier matching of points or objects.
    It is also beneficial to use 3D motion estimation in occluded/disoccluded regions since such regions do not have matching points, and the motion estimation is guided by the neighborhood motion only.
    Occluded regions typically belong to the relative background, and hence motion in such regions is similar to that of the neighborhood background.
    Estimating motion in 3D can utilize this correlation to estimate better object motion.
    
    We employ multi-plane images (MPI) as a 3D representation of the scenes, which represents the objects in the scene using multiple images placed at different depths.
    We choose the MPI representation since it can be directly processed by convolutional neural networks (CNN) and the frames can be reconstructed from MPI via differentiable alpha-compositing~\cite{srinivasan2019pushing}.
    We estimate 3D motion as displacement vectors between the corresponding points on the MPIs by training a CNN in an unsupervised fashion.
    Since MPI representations are inherently sparse, we process the MPIs using partial convolution layers and employ masked correlations to compute the 3D cost volumes.
    We feed the 3D cost volumes to the subsequent partial convolution layers, which estimate the displacement or flow vectors.
    Since the depth dimension in MPIs is discrete, we predict the motion in the depth dimension as a probability distribution over the depth planes.
    The expected value of this predicted distribution gives the displacement in the depth dimension.

    We then incorporate the available camera motion to determine all locations in the predicted frame that can be reconstructed from the past frame.
%    \textcolor{red}{Secondly, we generate the pixel intensities in the discoccluded regions of the predicted frame by adopting a flow field infilling approach to infill the backward flow vectors.
%    We hypothesize that the backward flow vectors that can be determined in regions that are visible in both the frames can help predict these vectors in the disoccluded regions.
%    In contrast, the disocclusion infilling in Gao \etal is based on a simple intensity infilling approach.}
    Employing a 3D infilling network similar to that of Srinivasan \etal~\cite{srinivasan2019pushing}, we synthesize the regions which are newly uncovered in the predicted frames.
    We dub our model as DeCOMPnet since we explicitly decompose the motion into camera and object motion for predicting the next frame.

    Since most view synthesis and video prediction datasets do not satisfy the problem assumptions for TVS-DS, we develop a new challenging dataset named the Indian Institute of Science Virtual Environment Exploration Dataset - Dynamic Scenes (IISc VEED-Dynamic).
    Our dataset contains 800 videos with 12 frames per video with a wide variety of camera and object motion.
    We render the videos using Blender at full HD resolution and a frame rate of 30fps.
    We evaluate our model and benchmark other video prediction and view synthesis models on our dataset and the MPI-Sintel~\cite{butler2012sintel} dataset for frame-rate upsampling.
    We show that our model achieves state-of-the-art performance in terms of the quality of the predicted frames.
    We further upper bound the performance of our model components using an oracle that has knowledge of the future frames.
    % We will publicly release our dataset and the code for our models.

    We summarize our main contributions as follows:
    \begin{itemize}
        \item We formulate a framework for temporal view synthesis of dynamic scenes that uses the available user or camera motion and only predicts the object motion.
        \item We design a 3D motion estimation model using an MPI representation of past frames after nullifying the camera motion between them.
        We introduce masked correlation and partial convolution layers to handle sparsity in the MPI representation.
        \item We develop a challenging dataset, IISc VEED-Dynamic, consisting of 800 videos at full HD resolution to evaluate our algorithm.
        We show that our model outperforms other competing models on both MPI-Sintel and our datasets.
    \end{itemize}

    \section{Related Work}\label{sec:related-work}

    \subsection{Video Prediction}\label{subsec:related-work-video-prediction}
    Deep video prediction was initially proposed as a self-supervised approach for representation learning of videos~\cite{srivastava2015unsupervised}.
    Video prediction has also found diverse applications such as robotic path planning~\cite{finn2016unsupervised}, anomaly detection~\cite{liu2018future}, video compression~\cite{liu2021deep} and autonomous driving~\cite{lotter2017deep}.
    Various video prediction approaches include multiscale prediction~\cite{mathieu2016deep}, predictive coding~\cite{lotter2017deep}, decomposing video into motion and content~\cite{villegas2017mcnet,tulyakov2018mocogan}, decoupling motion of background and foreground objects~\cite{wu2020future}, decomposing motion into velocity and acceleration maps~\cite{sarkar2021decomposing}, action conditioned prediction~\cite{lee2018stochastic} and so on.
    DPG~\cite{gao2019disentangling}, which disentangles motion propagation and content generation, is closely related to our work.
    However, our approach differs in decomposing the motion into camera motion and object motion and estimating object motion in 3D using MPIs.
    % Our work can also be considered as an extension of DPG~\cite{gao2019disentangling} that separates motion and infilling, where we further decompose the overall motion into camera motion and objects' motion.
    % Sarkar \etal~\cite{sarkar2021decomposing} decompose motion into velocity and acceleration maps by extending MCnet~\cite{villegas2017mcnet}, whereas we explicitly decouple camera and object motion into sequential steps.

    To account for uncertainty of future in long term prediction, stochastic video prediction models~\cite{babaeizadeh2018stochastic,denton2018stochastic,villegas2019high} aim to predict multiple future motion-trajectories for a given past.
    % Since our problem formulation needs a single next frame prediction, we do not benchmark stochastic prediction models
%    However, for frame rate upsampling, we need the models to provide a single prediction.
%    Moreover, the uncertainty in one-step prediction at high frame rates is minimal, and hence we do not benchmark stochastic prediction models.
    A detailed review of video prediction models can be found in ~\cite{oprea2020review}.
    However, video prediction models, in general, do not use camera motion and depth available in temporal view synthesis.
    In contrast, temporal view synthesis deals with the question of how to use camera motion and only predict the local motion of objects.
    %To the best of our knowledge, ours is the first model to use camera motion and depth to predict the next frame.

    \begin{figure*}
        \centering
        \includegraphics[width=\linewidth]{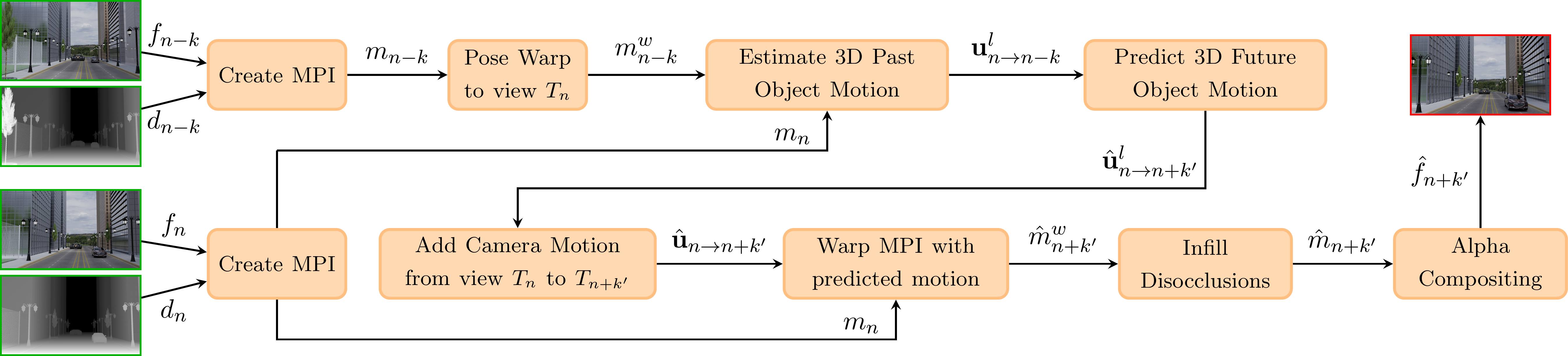}
        \caption{Overall architecture of DeCOMPnet.
        The given past frames are first converted to MPI and warped to the same camera view.
        3D object motion is estimated between the warped MPIs and extrapolated to predict the future object motion.
        Future camera motion is incorporated to predict the total future motion, which is used to warp the MPI of $f_n$.
        The warped MPI is then infilled and alpha composited to obtain the predicted future frame.
        For better visualization, inverse depth maps are shown.}
        \label{fig:architecture}
    \end{figure*}

    \subsection{View Synthesis}\label{subsec:related-work-view-synthesis}
    View Synthesis aims to synthesize the image as seen from a novel viewpoint given one or more images from different viewpoints.
    The models typically assume that camera pose is known, but depth is unknown and learn the depth either explicitly~\cite{wiles2020synsin,shih20203dp} using a depth estimation module or implicitly through representations such as MPI~\cite{zhou2018stereomag,srinivasan2019pushing}.
    Recently, neural radiance fields (NeRF)~\cite{mildenhall2020nerf} based models have found success in view synthesis.
    Depth image based rendering (DIBR) models assume depth is also known and typically employ the popular warp-and-infill approach and focus on infilling the disocclusions~\cite{cho2017hole,luo2020disocclusion}.
    Luo \etal\cite{luo2016hole} detect and remove foreground objects, reconstruct the background to infill the disocclusions and then apply motion compensation.
    Recently, Kanchana \etal~\cite{kanchana2022ivp} considered the problem of temporal view synthesis for static scenes.

    Different from novel view synthesis, dynamic view synthesis~\cite{gao2021dynamic,yoon2020novel,lin2021deep} aims to synthesize the video frames of a dynamic scene from a novel viewpoint.
    Different approaches include combining single-view and multi-view depth~\cite{yoon2020novel}, using static and dynamic NeRF~\cite{gao2021dynamic}, using MPI representations~\cite{lin2021deep,xing2021temporal} and so on.
    The above works assume that the scene is static between the views and do not predict any object motion.
    Recent works such as HyperNeRF~\cite{park2021hypernerf} and the model by Li \etal~\cite{li2021neural} generate frames between two time instants by interpolating in a higher dimensional hyper-space or using scene flow.
    However, these models need to be trained afresh when there is a change in the scene or on the arrival of new rendered frames, which may be infeasible in frame-rate upsampling.

    \subsection{Optical flow and scene flow}\label{subsec: related-work-flow}
    Optical flow estimation is a classical problem~\cite{lucas1981iterative,horn1981determining} which has found renewed interest due to the success of deep neural networks~\cite{sun2018pwcnet,liu2020arflow}.
    Optical flow methods estimate a dense field of displacement vectors in the 2D image plane by luminance constancy based matching of points between two frames.
    Scene flow~\cite{vedula1999three} extends optical flow to 3D\@.
    Recently, Yang \etal~\cite{yang2020upgrading} estimate scene flow by expanding 2D optical flow to 3D.
    % In particular, matching points in a pair of frames are unprojected to the 3D space and the scene flow is obtained as the distance between the unprojected points.
    Our 3D motion estimation differs from the above through the use of the 3D MPI representation.
    We find that estimating the object motion using a 3D representation achieves superior performance when compared to that using a 2D representation (\autoref{subsec:ablations}).

    \subsection{Image and Video Inpainting}\label{subsec:related-work-inpainting}
    In dynamic scenes, both the camera and object motion can create disocclusions in the next frame, which need to be infilled.
%    Thus we could use off-the-shelf classical ~\cite{criminisi2004region,barnes2009patchmatch,wexler2007space} or deep learning based ~\cite{pathak2016context,iizuka2017globally,yu2019free,nazeri2019edgeconnect,kim2019deep,xu2019deep,lee2019copy} image or video inpainting algorithms to infill the holes.
%    While the image and video inpainting primarily focus on arbitrary square or free-form holes, the holes in temporal view synthesis occur due to disocclusions.
%    Our disocclusion models differ in the way we exploit the motion that causes the disocclusions to infill the holes more effectively.
    Several image and video inpainting algorithms exist in the literature including classical~\cite{criminisi2004region,barnes2009patchmatch,wexler2007space} and deep learning~\cite{pathak2016context,iizuka2017globally,yu2019free,nazeri2019edgeconnect,kim2019deep,xu2019deep,lee2019copy} based models.
    The infilling model proposed by Srinivasan \etal~\cite{srinivasan2019pushing} exploits the 3D structure of the scene to infill disocclusions.

%     \subsection{Asynchronous Reprojection}\label{subsec:asynchronous-reprojection}
%     Asynchronous reprojection methods generate a next frame by reprojecting the latest rendered frame using the updated head position.
%     Asynchronous timewarp~\cite{oculus2016atw} generates the next frame by accounting for only the change in orientation of the camera.
%     Asynchronous spacewarp~\cite{oculus2016asw} additionally accounts for the object motion in the scene, and asynchronous spacewarp 2.0~\cite{oculus2019asw2} further extends it to account for translational motion of the camera.
% %    However, asynchronous spacewarp infills the disocclusions caused by translation motion of the camera by stretching the neighborhood~\cite{}.  % https://xinreality.com/wiki/Asynchronous_Spacewarp#:~:text=Asynchronous%20Spacewarp%20or%20ASW%20applies,run%20on%20lower%20performance%20hardware.
%     While our problem formulation is similar to asynchronous spacewarp, we explore a learning-based method.

    \section{Problem Statement}\label{sec:problem-statement}
    We formulate the problem of temporal view synthesis of dynamic scenes for causal frame rate upsampling of synthetic videos.
    Consider the scenario of upsampling by $k$ times, where we predict $k-1$ future frames before the next rendered frame.
    Given previous frames $ \left\{ f_{n},f_{n-k},\ldots, f_{n-lk} \right\} $, their depth maps $ \left\{ d_{n},d_{n-k},\ldots, d_{n-lk} \right\} $, camera poses (extrinsics) $ \left\{ T_{n},T_{n-k},\ldots, T_{n-lk} \right\} $, camera intrinsics $K$ and the camera poses of the next frames $ \left\{ T_{n+1}, T_{n+2}, \ldots, T_{n+k-1} \right\} $, we seek to predict the next frames $ \left\{ f_{n+1}, f_{n+2}, \ldots, f_{n+k-1} \right\} $.
    We assume that the motion in the video is caused by both camera and object motion.
    We refer to the motion due to user or camera movement as global motion and that of objects as local motion.

    Although the camera motion is available and large parts of the frame to be predicted can be generated by warping the previous frame to the desired view, the movement of objects creates additional challenges.
    An off-the-shelf application of video prediction algorithms can be inefficient since these algorithms do not effectively use the camera motion and the scene depth.
    Thus, the key challenge in predicting the next frame is to design a framework that can predict the motion of individual objects and utilize the available camera motion.
    % While camera motion can lead to the appearance of new regions into the frame, in VR applications, one can also render the frames at a slightly larger resolution and then crop out the new regions while displaying.
    % Thus, no new regions need to be synthesized during temporal view synthesis.
    We assume that the ground truth depth maps are available for the rendered frames since we focus on graphical rendering applications in this work.
%    Although recent graphical renderers can also provide motion vectors, most of the current gaming applications do not provide such information~\cite{}.
%    Further, if the video is streamed from a server, sending motion vectors leads to additional transmission overhead, which may not be possible depending on available bandwidth~\cite{splitXR}.
    We also assume that illumination changes are minimal due to the high frame rates of the videos.

    \section{Method}\label{sec:method}
    \subsection{Multi-Plane Images (MPI)}\label{subsec:mpi}
    Before delving into the details of our model, we briefly discuss the MPI representation and its generation.
    The MPI representation introduced by Zhou \etal~\cite{zhou2018stereomag} expands a 2D RGB frame into a set of RGBA image planes, located at different depths.
    The alpha channel ($\alpha \in [0, 1]$) in each plane denotes occupancy of the scene at the corresponding depth.
    Utilizing the knowledge of depth, we create the MPI directly from the RGB-D image instead of estimating the MPI as is common in literature~\cite{zhou2018stereomag,tucker2020single,li2021mine}.
    For the given RGB-D image, we first sample $Z$ planes uniformly in inverse depth between the minimum and maximum depth of the scene.
    For every location $\mathbf{x}$, we set $\alpha=1$ at the plane nearest to the true depth of $\mathbf{x}$ and set $\alpha=0$ for the rest of the planes.
    Thus at each location $\mathbf{x}$, the $\alpha$ values across all the planes form a one-hot vector.
    We modify the MPI representation to contain true depth values in an additional channel along with RGBA\@.
    We denote the MPI representation of $f_n$ as $m_n = \left\{ c_n, d_n, \alpha_n \right\}$, where $c_n$, $d_n$ and $\alpha_n$ are the RGB, depth and alpha channels respectively.
    To warp an MPI to a different camera view, we employ reprojection and bilinear splatting~\cite{tulsiani2018layer,kanchana2022ivp} instead of inverse homography employed by Zhou \etal~\cite{zhou2018stereomag}.
    Finally, to render a 2D frame from an MPI, we use alpha compositing in back to front order using the standard over operation~\cite{zhou2018stereomag}.

    \begin{figure*}
        \centering
        \includegraphics[width=\linewidth]{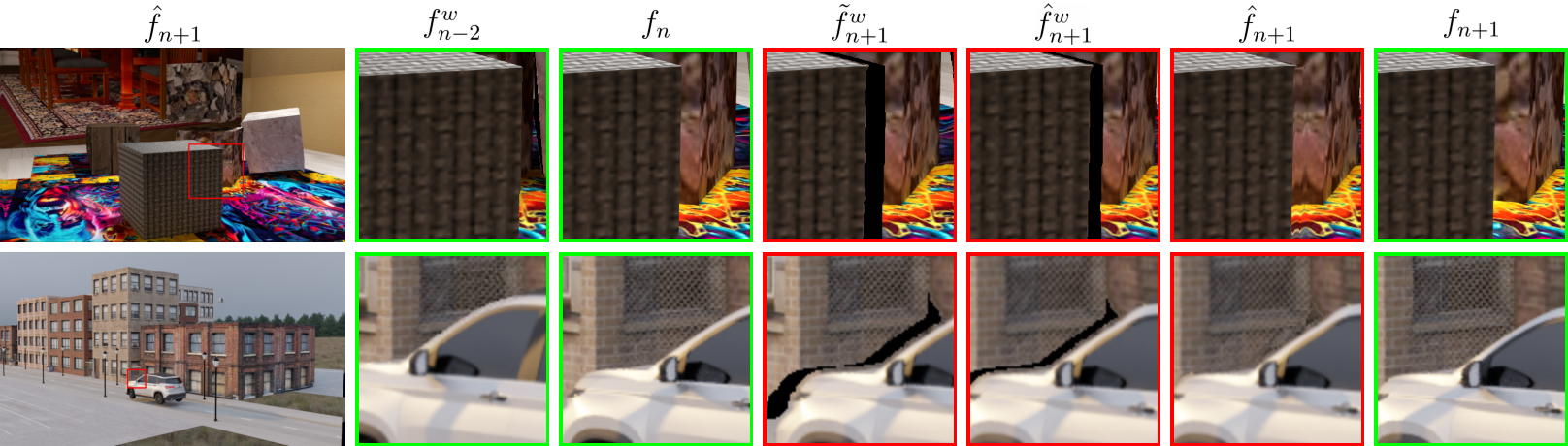}
        \caption{Visualization of outputs of various stages in our framework: Each row shows a different sample.
        The first column shows the full resolution frame and the subsequent columns show an enlarged region of a cropped region.
        The second and third columns show past frames after camera motion compensation.
        The fourth and fifth columns show the frame after predicting local and global motion respectively, which contain disocclusions (shown in black).
        The sixth column shows the result after infilling and the last column shows the true frame.
        }
        \label{fig:frame-lifecycle}
    \end{figure*}

    \subsection{Overview of the Proposed Approach}\label{subsec:overview}
    We present our approach for predicting future frames of dynamic scenes using camera motion knowledge as follows.
    For a scene with moving objects captured by a moving camera, to explicitly use the available camera motion in predicting the next frame, we adopt the following two-step approach.
    We first hold the camera still and account for the object motion.
    We then keep objects still and account for the camera motion alone.
    We use MPIs to represent the 3D scenes.
%    \textcolor{red}{Mathematically, let $m_n = \left\{ c^z_n, d^z_n, \alpha^z_n \right\}_{z=1}^{Z}$ denote the MPI representation of frame $f_n$, where $c^z_n$, $d^z_n$ and $\alpha^z_n$ are the RGB, depth and alpha channels respectively of a $Z$ plane MPI representation.}
    Let $k' \in \{1, 2, \ldots, k-1\}$ denote the prediction timestep and $\hat{\mathbf{u}}^{l}_{n \rightarrow n+k'}(\mathbf{x}, z)$ be the local optical flow in plane $z \in \left\{ 1, 2, \ldots, Z \right\}$ at location $\mathbf{x}$ that describes the motion of the pixel from time instant $n$ to $n+k'$ in view $T_n$.
    Let $P_{n \rightarrow n+k'}$ be the pose-warping operator from view $T_n$ to $T_{n+k'}$ that includes both the local and global motion in the warping.
    Corresponding to the location $(\mathbf{x}, z)$ in $m_n$ such that $\alpha_n(\mathbf{x}, z) = 1$, we obtain the MPI $\hat{m}^w_{n+k'}$ of a future frame $f_{n+k'}$ as
    \begin{align}
        \hat{m}^w_{n+k'}(P_{n \rightarrow n+k'}(\mathbf{x}, \hat{\mathbf{u}}^l_{n \rightarrow n+k'}(\mathbf{x}, z), d_n(\mathbf{x}, z))) = m_n(\mathbf{x}, z),
        \label{eq:total-motion-reconstruction}
    \end{align}
    where the pose-warping operator $P_{n \rightarrow n+k'}$ is defined as
    \begin{align}
        P_{n \rightarrow n+k'}(\mathbf{x}, \mathbf{u}, d) = K T_{n+k'} T_n^{-1} (d + \mathbf{u}_z) K^{-1} (\mathbf{x} + \mathbf{u}_{xy}),
        \label{eq:pose-warping-transformation}
    \end{align}
    where $\mathbf{u}_{xy}$ and $\mathbf{u}_z$ denote the components of flow in the x-y plane and in the depth dimension respectively.
%    We then reconstruct the frame $f_{n+1}$ from $m_{n+1}$ using alpha compositing~\cite{zhou2018stereomag}.
    Since \autoref{eq:pose-warping-transformation} represents forward warping, to obtain the intensities at integer locations of $\hat{m}^w_{n+k'}$, we use splatting similar to~\cite{tulsiani2018layer, kanchana2022ivp}.
    Along the depth dimension, we simply select the nearest plane.
    We omit the conversion between non-homogeneous and homogeneous coordinates for notation simplicity.

    In \autoref{eq:total-motion-reconstruction}, while the camera motion $P_{n \rightarrow n+k'}$ is known, the object motion $\hat{\mathbf{u}}^l_{n \rightarrow n+k'}$ is unknown and needs to be predicted.
    While warping $m_n$ to get $\hat{m}^w_{n+k'}$ using \autoref{eq:total-motion-reconstruction}, multiple locations from $m_n$ can map to a same location but different depth planes in $\hat{m}^w_{n+k'}$.
    Thus, for a few locations in $\hat{m}^w_{n+k'}$ across all the planes, there may be no matching points in $m_n$.
    Rendering such an MPI using alpha-compositing creates disocclusions or holes.
    Hence, we infill the warped MPI $\hat{m}^w_{n + k'}$ to get $\hat{m}_{n + k'}$ before rendering the frame $\hat{f}_{n+k'}$.
    We summarize our approach in \autoref{fig:architecture}.
%\textcolor{red}{, where we predict the 3D local motion $\hat{\mathbf{u}}^l_{n \rightarrow n+k'}$, use it along with the camera motion $P_{n \rightarrow n+k'}$ to warp $m_n$ to $m^w_{n+k'}$, infill the warped MPI to get $\hat{m}_{n + k'}$, and finally render it to get the predicted next frame $\hat{f}_{n+k'}$}.
    In the following subsections, we present the main challenges and our contributions in local motion prediction and briefly discuss our disocclusion infilling module.
    % Recall that our method mainly differs from DPG~\cite{gao2019disentangling} in the explicit use of the available camera motion and depth as well as disocclusion infilling for future frame prediction.

    \subsection{Local 3D Object Motion Prediction}\label{subsec:local-object-motion-prediction}
%    \begin{figure*}
%        \centering
%        \includegraphics[width=\linewidth]{res/images/ObjectMotionIsolation1}
%        \caption{\textbf{Object motion isolation:} The full frame $f_n$ is shown in the centre for context. The left column shows cropped regions of frames $f_n$ and $f_{n-2}$ and the right column shows cropped regions of frames $f_n$ and $f^w_{n-2}$. In this example, the chair is moving whereas the wall lamp is stationary. While this is not very apparent between frames $f_n$ and $f_{n-2}$, the object motion is clearly visible between the frames $f_n$ and $f^w_{n-2}$. Disoccluded regions in frame $f^w_{n-2}$ are shown in black. Best viewed in the attached video in the supplementary.}
%        \label{fig:object-motion-isolation}
%    \end{figure*}

    We predict the 3D object motion $\hat{\mathbf{u}}^l_{n \rightarrow n+k'}$ in view $T_n$ by estimating the local motion between $m_n$ and $m_{n-k}$ corresponding to $f_n$ and $f_{n-k}$, and extrapolating it.
    We only use the past ground truth frames to avoid the accumulation of errors.
    Since the motion between $f_n$ and $f_{n-k}$ is a mixture of both global and local motion, the local motion alone needs to be extracted from the overall motion.
    To achieve this, we first nullify the global motion between the past frames by warping $m_{n-k}$ from view $T_{n-k}$ to $T_n$ to get $m^w_{n-k}$, using \autoref{eq:pose-warping-transformation} by setting $\mathbf{u}=0$.
%    Although one could create MPIs $m_n, m_{n-2}$ first and then warp $m_{n-2}$, for efficient caching of training data, we first warp $f_{n-2}$ to the view $T_n$ to get $f^w_{n-2}$ and then create the MPI $m^w_{n-2}$ as described in \autoref{subsec:mpi}.
    Thus, the residual motion between $m_n$ and $m^w_{n-k}$ corresponds to the object motion between time instants $n$ and $n-k$.
    We estimate the 3D optical flow between $m_n$ and $m^w_{n-k}$ to compute this local motion and use it to predict $\hat{\mathbf{u}}^l_{n \rightarrow n+k'}$.

    \textbf{Past 3D flow estimation:}
    Given the success of deep convolutional neural networks for optical flow estimation, we explore such an approach to estimate the flow between the MPI representations.
    We encounter two challenges while estimating object motion between two MPIs.
    The first is that MPI representations are inherently sparse, \ie a significant number of pixels in MPIs have $\alpha=0$.
    We handle the sparsity of MPIs by introducing 3D partial convolution layers, which convolve the input only in the regions where $\alpha=1$.
    2D partial convolutions were introduced by Liu \etal~\cite{liu2018pconv} to infill holes in image inpainting applications.
    However, we apply partial convolution in a completely different domain of estimating 3D optical flow with MPIs.
    In this regard, we modify the partial convolution layer to not dilate the alpha mask at every layer, since our work aims to estimate optical flow where $\alpha=1$.
    Estimating optical flow typically requires computing a cost volume using a correlation layer~\cite{sun2018pwcnet}.
    We design masked correlation layers to handle the sparsity of MPI while computing the 3D cost volume.
    For input features $\mathbf{h}_1, \mathbf{h}_2$ along with corresponding alpha masks $\alpha_{h_1}, \alpha_{h_2}$, we compute the cost volume and the corresponding alpha mask as
    \begin{align}
        \nonumber \text{cv}((\mathbf{x}_1, z_1),(\mathbf{x}_2, z_2)) = (\mathbf{h}_1 &(\mathbf{x}_1, z_1) \alpha_{h_1}(\mathbf{x}_1, z_1))^T \cdot  \\ &(\mathbf{h}_2(\mathbf{x}_2, z_2) \alpha_{h_2}(\mathbf{x}_2, z_2)), \label{eq:masked-correlation}\\
    % \end{align}
    % and the corresponding alpha mask as
    % \begin{align}
        \alpha_{\text{cv}}((\mathbf{x}_1, z_2),(\mathbf{x}_2, z_2)) = \alpha_{h_1}(&\mathbf{x}_1, z_2) \cdot \alpha_{h_2}(\mathbf{x}_2, z_2).  \label{eq:masked-correlation-alpha}
    \end{align}
    The above cost volume and mask are then fed to subsequent partial convolution layers to estimate the optical flow.

    The second challenge is in representing the 3D flow due to the discrete nature of depth planes in the MPI representation.
    We use real-valued displacements $\mathbf{a} \in \mathbb{R}^2$ in the x-y dimensions.
    In the depth dimension, we model the flow at location $(\mathbf{x}, z)$ as a difference in the index of the planes in the MPI representation, from $m_n$ to $m^w_{n-k}$.
    % to one of $2 s_z + 1$ adjacent planes of $(\mathbf{x}, z)$.
    We refer to this difference as $z'$, where $z' \in \left\{ -s_z, -s_z+1, \ldots, 0, \ldots, s_z-1, s_z \right\}$ and $2 s_z + 1$ is the size of the window around the plane $z$ in the depth dimension.
    The network outputs a probability distribution $b_{z'}$ on the differences $z'$.
    Thus,
    % \[\mathbf{v}^z(\mathbf{x}, z) \in [0, 1]^{2 s_z + 1} : \Vert {\mathbf{v}^z(\mathbf{x}, z)} \Vert_1 = 1 \ \ \forall \mathbf{x}.\]
    \begin{align}
      b_{z'}(\mathbf{x}, z) \in [0, 1] : \sum_{z'=-s_z}^{s_z} {b_{z'}(\mathbf{x}, z)} = 1 \ \ \forall (\mathbf{x}, z).  
    \end{align}
    % \[b_{z'}(\mathbf{x}, z) \in [0, 1] : \sum_{z'=-s_z}^{s_z} {b_{z'}(\mathbf{x}, z)} = 1 \ \ \forall (\mathbf{x}, z). \]
%    Mathematically, the output of the deep neural network is a field of flow vectors, $\tilde{\mathbf{u}} \in \mathbb{R}^{2 + s_z}$ where $\tilde{\mathbf{u}}=(\tilde{\mathbf{u}}^{xy}, \tilde{\mathbf{u}}^z)$ with $\tilde{\mathbf{u}}^{xy} \in \mathbb{R}^2$ and $\tilde{\mathbf{u}}^z \in \mathbb{R}^{s_z}, \ \sum_{z}{\tilde{\mathbf{u}}^z} \left( \mathbf{x}, z \right) = 1 \ \ \forall \mathbf{x} $.
%    Here, $\tilde{\mathbf{u}}^z$ is estimated across the $s_z$ neighboring depth planes used to compute the cost volume.
    Implementing \autoref{eq:total-motion-reconstruction} requires a real-valued 3D flow vector, $\mathbf{u}^l_{n \rightarrow n-k}$, which we compute as
    \begin{align}
        \nonumber \mathbf{u}^l_{n \rightarrow n-k}\left(\mathbf{x}, z\right) = \Biggl( \mathbf{a}\left(\mathbf{x}, z\right), \left( \sum_{z'=-s_z}^{s_z} {b_{z'}\left(\mathbf{x}, z\right)} d\left(z + z'\right)\right)  & - d(z) \Biggr) \\
        & \in \mathbb{R}^3,
        \label{eq:flow-quantized-to-real}
    \end{align}
    where $d\left(z\right)$ is the depth corresponding to the $z^{\text{th}}$ plane in the MPI\@.
%    The second term in \autoref{eq:flow-quantized-to-real} computes the displacement of point at $(\mathbf{x}, z)$ using the expected depth of the point after motion.
%    For brevity, we do not show the dependence of $\mathbf{v}^z$ on $z'$ in \autoref{eq:flow-quantized-to-real}.

%    \textcolor{blue}{We estimate the motion in multiscale, in which we estimate the motion at a lower scale and then estimate and add the residual motion at higher scales.
%    However, we cannot directly add the predicted probability distributions.
%    Instead, we compute the expectation of the previous stage flow distribution with respect to the distribution predicted by the next stage (refer to supplement for mathematical details).}

    \begin{figure*}
        \centering
        \includegraphics[width=\linewidth]{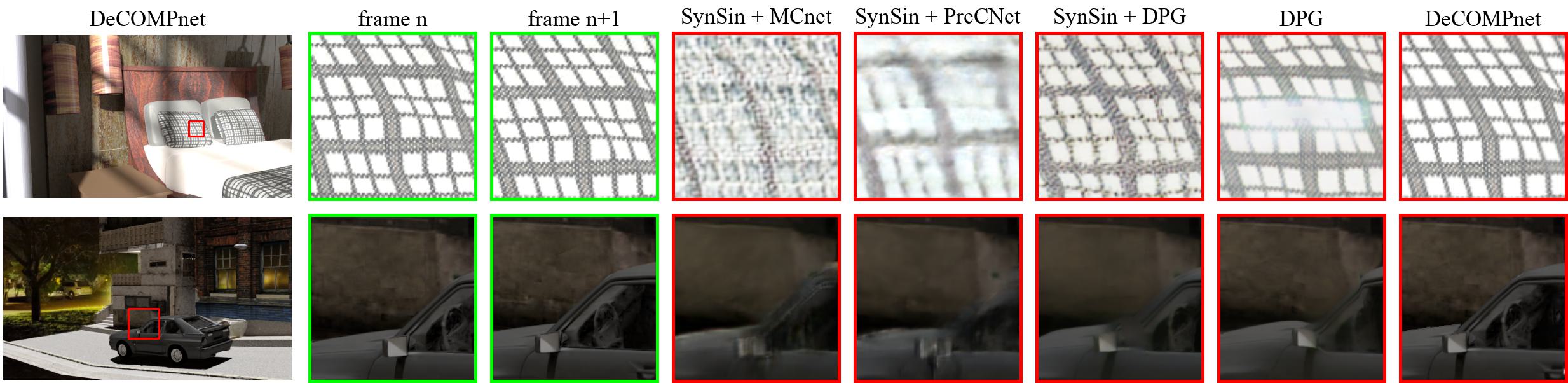}
        \caption{Qualitative comparisons on our dataset for single frame prediction.
        The first column shows a predicted frame by our model, DeCOMPnet, and the subsequent columns show enlarged versions of a cropped region for different models.
        The frames with green border are graphically rendered, and those with red border are predicted by different models.
%        In the scene in the first row, the books are moving up, and in the second scene, the pillows along with the bed are moving towards the camera.
        In the scene in the first row, the pillows along with the bed are moving towards the camera.
        The car is moving left in the second scene.
        All scenes have camera motion in addition to object motion.
        We observe that other models fail to produce sharp predictions or retain the object shape, whereas our model has retained the shape and textures.
        }
        \label{fig:qualitative-comparisons-ours}
    \end{figure*}

    We incorporate the above and design a multi-scale 3D flow estimation network using PWC-Net~\cite{sun2018pwcnet} as the backbone architecture.
    PWC-Net consists of an encoder-decoder style architecture, where optical flow is estimated in a coarse-to-fine manner.
    Specifically, we first obtain multi-scale 3D features of $m_n$ and $m^w_{n-k}$ using encoders at each scale with 3D partial convolutions and downsampling layers consisting of strided convolutions.
    Since the number of MPI planes is much smaller than the resolution of the other two spatial dimensions, we do not downsample/upsample the features along the depth dimension.
    At the decoder in each scale except the lowest one, we upsample the flow estimated by the previous scale.
    Using this flow, we warp the features of $m^w_{n-k}$ and feed it to the masked correlation layers, along with the features of $m_n$.
    The masked correlation layers, as described in \autoref{eq:masked-correlation} and \autoref{eq:masked-correlation-alpha}, output a cost volume which is then processed by subsequent partial convolution layers to estimate the residual flow at that scale.
    We estimate the final flow at two scales lower than the original resolution and upsample it by four times, as is popular in deep flow estimation models~\cite{sun2018pwcnet}.
    Network details are in the supplementary.
    We train the optical flow network $\mathcal{F}_\Theta$, with trainable parameters $\Theta$ to estimate the flow from $m_n$ to $m^w_{n-k}$ as
    \begin{align}
        \mathbf{u}^l_{n \rightarrow n-k} = \mathcal{F}_\Theta(m_n, m^w_{n-k}).
    \end{align}

    \textbf{Loss functions:}
    We train the network $\mathcal{F}_{\Theta}$ in an unsupervised fashion with a linear combination of photometric loss $\mathcal{L}_{\text{ph}}$ and a smoothness loss $\mathcal{L}_{\text{smooth}}$.
    Specifically, we warp $m^w_{n-k}$ using $\mathbf{u}^l_{n \rightarrow n-k}$ to reconstruct $\hat{m}_n$.
    % We use a combination of mean absolute error (MAE) and structural similarity (SSIM)~\cite{wang2004image} for the photometric loss as
    Photometric loss is a combination of mean absolute error (MAE) and structural similarity (SSIM)~\cite{wang2004image} as
    \begin{align}
        \mathcal{L}_{\text{ph}} = \beta \Vert (m_n - \hat{m}_n)\odot o_n \Vert_1 + (1 - \beta) \frac{1 - \text{\small{SSIM}}(m_n \odot o_n, \hat{m}_n \odot o_n)}{2},
        \label{eq:loss-optical-flow-photometric}
    \end{align}
    where, $\beta$ is a scaling constant, $o_n$ is the occlusion mask and $\odot$ represents the element-wise product.
    The MAE and SSIM losses are computed in each of the $Z$ planes and averaged.

    Unsupervised optical flow algorithms~\cite{meister2018unflow} compute photometric loss in the non-occluded regions only using an occlusion mask $o_n$ as in \autoref{eq:loss-optical-flow-photometric}.
    The occlusion mask is typically computed using forward-backward consistency of the optical flow.
    We instead utilize the 3D representation of the scene and determine the occluded pixels as those which are hidden after warping $m_n$ with $\mathbf{u}^l_{n \rightarrow n-k}$.
    Mathematically, we forward-warp $m_n$ using $\mathbf{u}^l_{n \rightarrow n-k}$ to get $\hat{m}_{n-k}$.
    We compute a visibility mask for $\hat{m}_{n-k}$ as
    \begin{align}
        \hat{v}_{n-k}(\mathbf{x}, z) = \prod_{y=1}^{z-1}(1 - \hat{\alpha}_{n-k}(\mathbf{x}, y)).
    \end{align}
    We then backward-warp $\hat{v}_{n-k}$ using $\mathbf{u}^l_{n \rightarrow n-k}$ to get $\hat{v}_n$.
    Finally, we compute the occlusion mask as
    \begin{align}
       o_n = \mathbbm{1}_{\left\{ \hat{v}_n > 0.5 \right\}}.
    \end{align}
    A value of $0$ in $o_n$ indicates that the point is occluded.
%    To help the network utilize the input depth in flow estimation, we also impose MAE loss on depth $\mathcal{L}_{\text{depth}}$ between $d_n$ and the warped depth $\hat{d}_n$ obtained similar to $\hat{f}_n$ as
%    \begin{align}
%        \mathcal{L}_{\text{depth}} = & \Vert d_n - \hat{d}_n \Vert_1 .
%    \end{align}
    For the edge-aware smoothness loss, along with gradients of RGB, we also use gradients of alpha channel to weigh the smoothness term as
    \begin{align}
        \mathcal{L}_{\text{smooth}} = (1 - \nabla \alpha_n) \cdot \exp(-a \cdot \nabla c_n) \cdot \nabla \mathbf{u}^l_{n \rightarrow n-k},
    \end{align}
where $a$ is a scaling constant.
    Thus, our overall loss function is
    \begin{align}
        \mathcal{L}_{\text{of}} = \mathcal{L}_{\text{ph}} + \lambda \mathcal{L}_{\text{smooth}}.
    \end{align}

    \textbf{Future flow prediction:}
    We employ a linear motion model~\cite{bao2019memcnet,bao2019dain} to predict the future flow as
    \begin{align}
        \hat{\mathbf{u}}^l_{n \rightarrow n+k'}(\mathbf{x}, z) = - \frac{k'}{k} \ \mathbf{u}^l_{n \rightarrow n-k}(\mathbf{x}, z).
        \label{eq:flow-prediction}
    \end{align}
    Thus, to predict the future local motion, we first isolate the local motion between the past frames by nullifying the global motion between them and then estimate the local motion as 3D optical flow between the MPIs of the past frames.
    We then extrapolate the past motion to predict the future motion.
%    We incorporate the above predicted future local motion with the available camera motion and warp $f_n$ to $\hat{f}^w_{n+1}$ following \autoref{eq:total-motion-reconstruction}.

    \subsection{Disocclusion Infilling}\label{subsec:infilling}
    As argued earlier, implementing \autoref{eq:total-motion-reconstruction} creates disocclusions.
    Hence we infill the disoccluded regions in $\hat{m}^w_{n+k'}$ using an approach similar to the one used by Srinivasan \etal~\cite{srinivasan2019pushing}.
    We feed $\hat{m}^w_{n+k'}$ to a 3D U-Net and predict 2D infilling vectors in the disoccluded regions that point to known regions in the same plane of MPI\@.
    We then infill the disoccluded regions by copying the intensities and alpha from the locations pointed by the predicted infilling vectors to obtain $\hat{m}_{n+k'}$.
    Alpha-compositing $\hat{m}_{n+k'}$ generates the predicted frame $\hat{f}_{n+k'}$.
    We train the disocclusion infilling network with mean squared error loss between the predicted frame $\hat{f}_{n+k'}$ and the true frame $f_{n+k'}$.
    We find that the network fails to completely infill large disoccluded regions, leaving partially unfilled disoccluded regions.
    Hence, during inference, we iteratively infill the disoccluded regions $g$ times by recursively feeding the infilled MPI to the network.

    \section{Experiments}\label{sec:experiments}

    \begin{figure*}
        \centering
        \includegraphics[width=\linewidth]{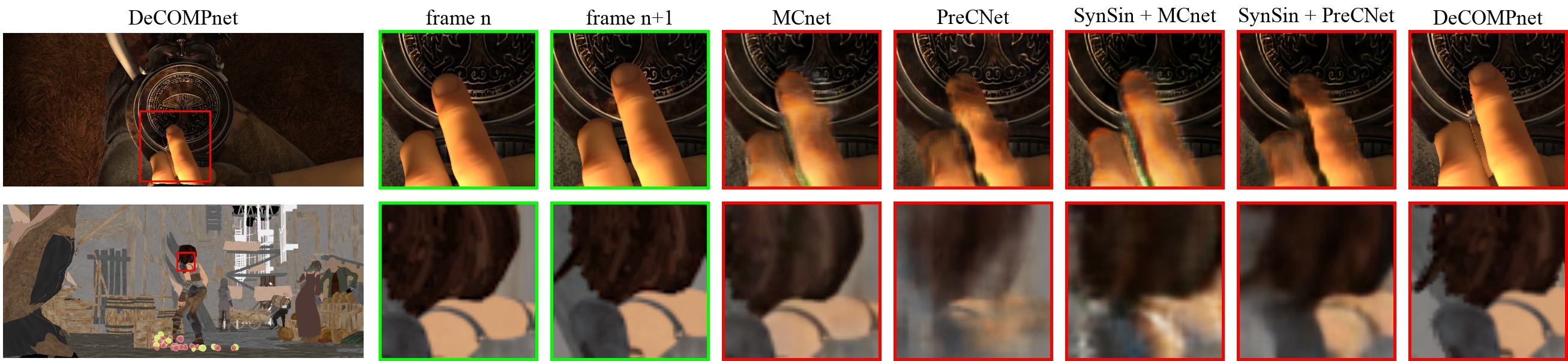}
        \caption{Qualitative comparisons on the MPI Sintel dataset for single frame prediction.
        The fingers are moving up in the first scene, and the girl is moving to the right in the second scene.
        We observe that our model has retained the shape of the objects, which the other models fail to.
        % MCnet has failed to predict any motion.
        }
        \label{fig:qualitative-comparisons-sintel}
    \end{figure*}

    \begin{table*}
        \centering
        \setlength\tabcolsep{8pt}
        \caption{Quantitative comparison of different models on ours and MPI Sintel datasets for single frame prediction. Models indicated with $^*$ are a combination of view synthesis and video prediction models, that we design.}
        \begin{tabular}{l|cccc|cccc}
            \hline
             & \multicolumn{4}{c|}{Our Dataset} & \multicolumn{4} {c}{MPI Sintel} \\
            \textbf{Model} & \textbf{PSNR \textuparrow} & \textbf{SSIM \textuparrow} & \textbf{LPIPS \textdownarrow} & \textbf{ST-RRED \textdownarrow} & \textbf{PSNR \textuparrow} & \textbf{SSIM \textuparrow} & \textbf{LPIPS \textdownarrow} & \textbf{ST-RRED \textdownarrow} \\
            \hline
            MCnet~\cite{villegas2017mcnet} & 24.66 & 0.7813 & 0.2406 & 207 & 24.00 & 0.7511 & 0.2230 & 530\\
            DPG~\cite{gao2019disentangling} & 28.24 & 0.8634 & 0.1091 & 71 & 20.00 & 0.6385 & 0.3056 & 1129\\
            PreCNet~\cite{straka2020precnet} & 24.86 & 0.8191 & 0.2409 & 244 & 25.60 & 0.7952 & 0.2463 & 571\\
            SynSin~\cite{wiles2020synsin} + MCnet$^*$ & 26.87 & 0.8254 & 0.1567 & 92 & 25.67 & 0.8031 & 0.1639 & 315\\
            SynSin + DPG$^*$ & 27.30 & 0.8461 & 0.1268 & 74 & 23.77 & 0.7795 & 0.2520 & 600 \\
            SynSin + PreCNet$^*$ & 26.81 & 0.8432 & 0.1508 & 100 & 25.92 & 0.8205 & 0.1581 & 330 \\
            \textbf{DeCOMPnet} & \textbf{30.60} & \textbf{0.9314} & \textbf{0.0634} & \textbf{28} & \textbf{29.64} & \textbf{0.8975} & \textbf{0.1032} & \textbf{259} \\
            \hline
        \end{tabular}
        \label{tab:results-ourdb-sintel}
    \end{table*}

    \subsection{Datasets}\label{subsec:datasets}
    \textbf{Our Dataset}: We develop a new dataset of videos with both camera and object motion due to the lack of any large scale datasets suitable for evaluating temporal view synthesis of dynamic scenes.
    We render the videos of our dataset with Blender using blend files from blendswap~\cite{blendswap2021blendswap} and turbosquid~\cite{turbosquid2021turbosquid} and add camera and object motion to the scenes.
    We add motion to the pre-existing scene objects or add new objects to the scene and animate them.
    Our dataset contains 200 diverse scenes of indoor environments such as hospital, kitchen, restaurant, and supermarket and outdoor environments like village, poolside, street, lake and so on.
    The scenes contain various moving objects such as books, chairs, tables, cars, airplanes, etc.
    For every scene, we generate four different camera trajectories covering different parts of the scenes and different kinds of object motion.
    Each sequence has 12 frames rendered at full HD resolution ($1920 \times 1080$) and 30fps.
    Thus, our dataset consists of 800 videos with 9600 frames in total.
    For every frame in our dataset, we store the corresponding ground truth depth, camera pose, and camera intrinsics.
    We use 135 scenes for training and 65 for testing.

    \textbf{MPI-Sintel}: The MPI-Sintel dataset~\cite{butler2012sintel}, which is widely used for evaluating optical flow estimation algorithms, contains both camera and object movement and also provides the ground truth depth and camera poses.
    Thus, it can be used to evaluate temporal view synthesis models.
    Since the required ground truth is provided for the train set only, we further divide the train set into train and test sets.
    The videos have a resolution of $1024 \times 436$ at $24$ frames per second.
    We use 13 scenes for training and 10 scenes for testing.
%    However, the dataset has only xx scenes for training and xx scenes for testing and is not designed to bring out the challenges in temporal view synthesis.

    We experiment on synthetic datasets only and not on real world datasets since our problem formulation is motivated by use-cases in increasing the frame rate for graphical rendering.
    Thus, we assume that the depth is available.

    \subsection{Comparisons}\label{subsec:comparisons}
    We compare our model against a combination of video prediction and view synthesis models.
    We use MCnet~\cite{villegas2017mcnet}, a popular video prediction model, PreCNet~\cite{straka2020precnet}, a recent model based on predictive coding and DPG~\cite{gao2019disentangling}, a model based on flow prediction and disocclusion infilling.
    For all the models, we use four past frames.
    Therefore, the prediction of first few frames uses the true past frames and the subsequent predictions use the previously predicted frames.

    Since the above methods do not make use of camera motion, we combine these video prediction models with a recent view synthesis model, SynSin~\cite{wiles2020synsin}.
    We first incorporate the camera motion by warping the past frames $f_{n}$, $f_{n-1}$, $f_{n-2}$ and $f_{n-3}$ to the view of $f_{n+1}$ using SynSin.
    We use the ablation model of SynSin, which uses true depth of the past frames.
    We then use video prediction models such as MCnet, DPG, or PreCNet on these warped frames to account for local motion and predict the desired frame.
    In order to guage the performance capability of this approach, we feed the warped $f_{n-1}$ and $f_{n-3}$ to the video prediction model, although these are not available during frame rate upsampling.

    We implement DPG ourselves and train the model on $256 \times 256$ patches on both datasets.
    For MCnet, PreCNet, and SynSin, we use the code and pretrained models provided by the authors and finetune them on both datasets.
    We test both the pretrained and the finetuned models and report the best performance.
    
    \begin{figure*}
        \centering
        \includegraphics[width=\linewidth]{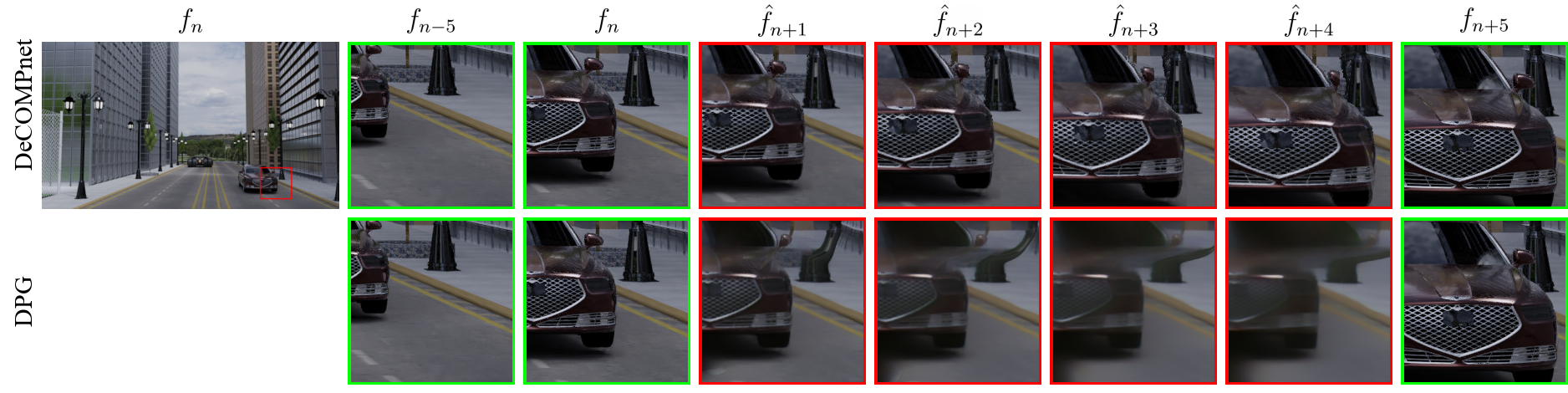}
        \caption{Multi frame predictions by DeCOMPnet.
        The first column shows $f_n$ at full resolution and the subsequent columns focus on a cropped region of $f_{n-5}, f_n$ and the four predicted frames.
        The last column shows $f_{n+5}$ for reference.
        }
        \label{fig:multistep-qualitative}
    \end{figure*}
    \begin{figure*}
        \centering
        \begin{subfigure}{.24\textwidth}
            \centering
            \includegraphics[width=0.95\linewidth]{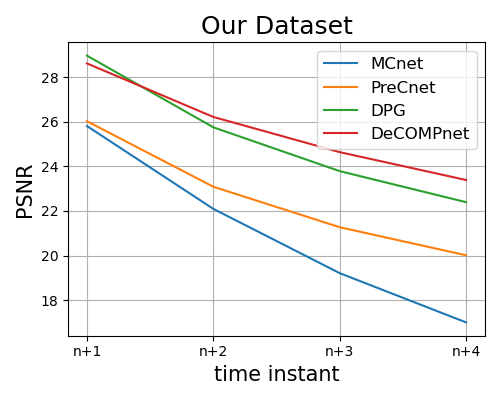}
        \end{subfigure}
        \begin{subfigure}{.24\textwidth}
            \centering
            \includegraphics[width=0.95\linewidth]{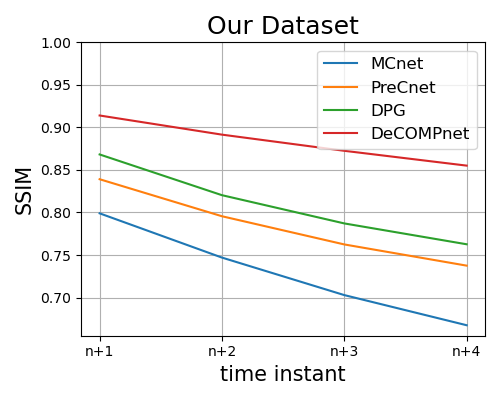}
        \end{subfigure}
        \begin{subfigure}{.24\textwidth}
            \centering
            \includegraphics[width=0.95\linewidth]{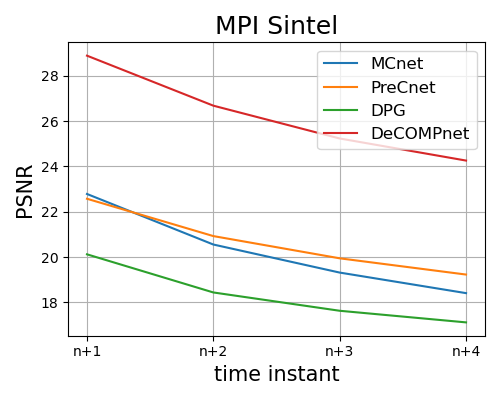}
        \end{subfigure}
        \begin{subfigure}{.24\textwidth}
            \centering
            \includegraphics[width=0.95\linewidth]{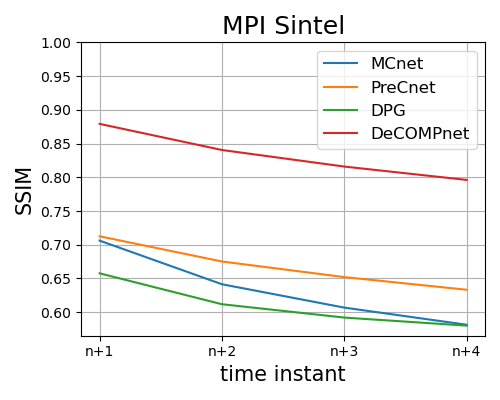}
        \end{subfigure}
        \caption{Quantitative comparison of the proposed DeCOMPnet against competing methods for multi frame prediction.
%        The two plots on left show PSNR and SSIM scores on Our dataset and the two on the right show the scores on the MPI Sintel Dataset.
        The plots show average quality score for the predicted frames $\hat{f}_{n+1}, \hat{f}_{n+2}, \hat{f}_{n+3}$ and $\hat{f}_{n+4}$.
        }
        \label{fig:multistep-quantitative}
    \end{figure*}

    \textbf{Implementation details:}
    We train the optical flow estimation network $\mathcal{F}_{\Theta}$ and the disocclusion infilling network separately due to GPU memory constraints.
    We initialize our flow estimation network using pretrained weights provided by ARFlow~\cite{liu2020arflow} and finetune it on the respective datasets.
    We modify the pretrained weights appropriately to work for 3D convolutions.
    We train both the networks for $10000$ iterations with patches of size $256 \times 256$ and a batch size of $4$.
    % We use a single NVIDIA RTX 2080 Ti GPU for all our experiments.
    Please refer to supplementary for the architecture details of the networks.
    We set the hyper-parameters as $s_z=1, Z = 4, \beta = 0.15, a = 10, \lambda=10, g=3$.
%    The optical flow and disocclusion infilling networks take about 13h and 25h to train respectively on our dataset.

    \textbf{Evaluation Measures:}
    We evaluate the predicted frames using various image quality measures such as peak signal-to-noise ratio (PSNR), structural similarity index (SSIM)~\cite{wang2004image} and LPIPS~\cite{zhang2018unreasonable}.
    Further, since image quality measures do not evaluate temporal quality, we also employ a video quality assessment measure, ST-RRED~\cite{soundararajan2013video} that measures both the spatial and temporal quality of the predicted frames.
    Since the focus of this work is not on predicting new regions entering the scene, we crop out 40 pixels on the top and bottom of the frames and 60 pixels on the left and right sides of the frames before evaluating the predictions.
    \begin{table}
        \centering
        \caption{Comparison of average endpoint error for the flows predicted by different ablated models, for single frame prediction.}
        \begin{tabular}{l|c}
            \hline
            \textbf{Model} & \textbf{Endpoint Error} \\
            \hline
            2D Flow & 2.8 \\
            3D Flow w/o p-conv and mask-corr & 2.0 \\
            \textbf{3D Flow} & \textbf{1.7} \\
            \hline
        \end{tabular}
        \label{tab:ablations}
    \end{table}

    \subsection{Single Frame Prediction}\label{subsec:single-frame-prediction}
    In single frame prediction, the goal is to predict every alternate frame and this can be studied by setting $k=2$ in our problem definition.
    Specifically, to predict $\hat{f}_{n+1}$, we use $f_{n}$ and $f_{n-2}$.
    % For the other competing video prediction benchmarks, we use $f_{n}, f_{n-1}, f_{n-2}$ and $f_{n-3}$ to predict $\hat{f}_{n+1}$.

    We first present examples of a few future frame predictions by DeCOMPnet and visualizations of outputs of various stages in our framework in \autoref{fig:frame-lifecycle}.
    In particular, we show the outputs after predicting the object motion alone, $\tilde{f}^w_{n+1}$, and after incorporating the global motion.
    Since such outputs are in the MPI representation space, we use alpha-compositing to obtain the corresponding images.

    We compare the quantitative results of DeCOMPnet against the competing methods in \autoref{tab:results-ourdb-sintel}.
    Our model outperforms all the competing methods in terms of all the quality measures.
    The relatively lower ST-RRED scores for DeCOMPnet indicate that the predictions by our model are superior in temporal quality.
    We observe that most models perform better on our dataset than on the MPI-Sintel dataset in general.
    This may be because the MPI-Sintel dataset has complex motion to make it challenging for optical flow estimation, making it even more challenging for prediction.
    We also observe that combining SynSin with video prediction models improves their performances, except for DPG on our dataset.
    Since DPG is performing reasonably well, when combined with SynSin, the artifacts introduced by SynSin may lead to a decrease in performance.
    However, on the MPI-Sintel dataset, since the performance of DPG is lower, it benefits from using SynSin.
    % We observe that using view synthesis with video prediction does better than video prediction alone.
    Further, we note that even though the combination of view synthesis and video prediction models use the knowledge of the true frames $f_{n-1}$ and $f_{n-3}$ which are not available at test time, our model still shows superior performance.

    \begin{figure*}
        \centering
        \includegraphics[width=\linewidth]{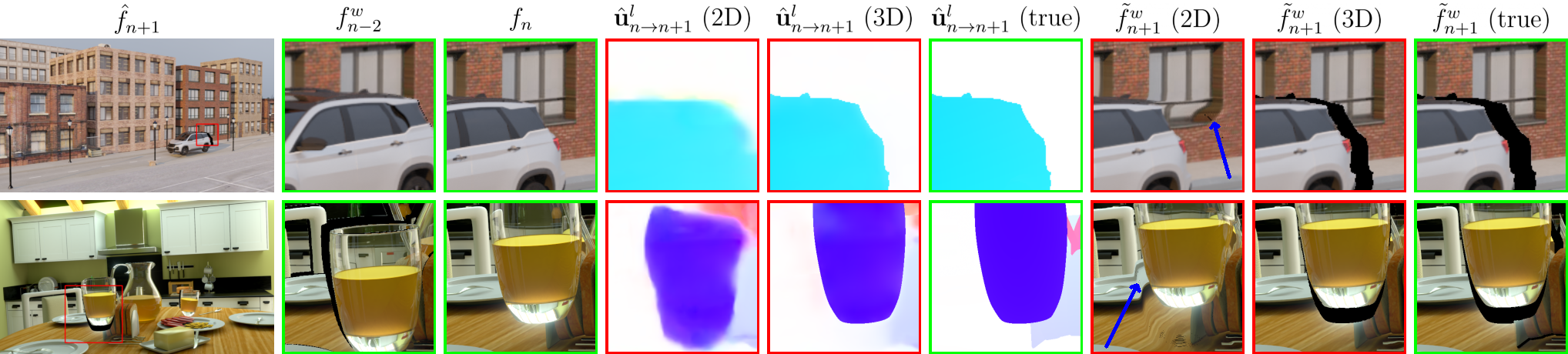}
        \caption{Qualitative comparison of 2D and 3D flow estimations for single frame prediction.
%        The first column shows the full scene for context and the subsequent columns show an enlarged region from the frame.
        The second and third columns show the input frames $f_n$, $f^w_{n-2}$ to the flow estimation networks. Fourth, fifth and sixth columns visualize the x-y component of flows predicted by the models and the ground truth flow.
        The next three columns show the corresponding frames $\tilde{f}^w_{n+1}$ reconstructed by applying local flow $\hat{\mathbf{u}}^l_{n \rightarrow n+1}$ on $f_n$.
        Notice the sharpness of 3D flow and the distortions in the background of the frame reconstructed with 2D flow as pointed by the blue arrow.
        Disoccluded regions are shown in black.
        Here we only visualize object motion prediction and do not show the final predicted frame.
        Global motion and infilling need to be applied on top of $\tilde{f}^w_{n+1}$ as shown in \autoref{fig:frame-lifecycle} to obtain $\hat{f}_{n+1}$.
        Optical flow visualization is similar to Baker \etal~\cite{baker2011middlebury}.
%        We observe that, the model without depth distorts the shape of the basket in the predicted frame, whereas our model with depth is able to maintain the shape of the basket.
        }
        \label{fig:flow-2d-vs-3d}
    \end{figure*}

    We show the qualitative results of our model and the benchmarked models in~\autoref{fig:qualitative-comparisons-ours} and ~\autoref{fig:qualitative-comparisons-sintel}.
    While other models introduce artifacts such as blur or distortions in the shape and texture of objects, DeCOMPnet predicts the future frame reasonably well.
%    In particular, we believe that the use of depth in the flow prediction helps our model avoid distortions in object shapes as compared to DPG\@.
    To notice the temporal superiority of our model, view supplementary videos.
%    The videos show that our model can predict object motion significantly better than the competing methods.
%    \textcolor{blue}{We believe this is due to our robust estimation of object motion in 3D using MPI representation and the decoupling of the camera and object motion}.

    \subsection{Multi Frame Prediction}\label{subsec:multi-frame-prediction}
    We now analyze the ability of our model to predict multiple frames into the future.
    In particular, we study frame-rate upsampling by a factor of five times by setting $k=5$.
%    We do not observe performance improvements by re-training our model for multi-step prediction, and hence we use the model trained for single frame prediction.
    In our framework, we estimate the object motion $\mathbf{u}^l_{n \rightarrow {n-5}}$ only once, and compute the predicted motion for each of the future time steps using \autoref{eq:flow-prediction}.
    We then use \autoref{eq:total-motion-reconstruction} to warp $m_n$ to $\hat{m}^w_{n+1}, \hat{m}^w_{n+2}, \hat{m}^w_{n+3}$ and $\hat{m}^w_{n+4}$, which are then infilled and alpha composited to predict the future frames.
    For the benchmark comparison models, $f_{n}, f_{n-1}, f_{n-2}$ and $f_{n-3}$ are used to predict $\hat{f}_{n+1}$.
%    Further, $\hat{f}_{n+1}, f_{n}, f_{n-1}$ and $f_{n-2}$ are used to predict $\hat{f}_{n+2}$, and so on.
    Thus, compared to our model, the benchmarked models have the additional knowledge of $f_{n-1}, f_{n-2}$ and $f_{n-3}$.
    Although, these frames are not available in practice, the goal of this experiment is to analyze the performance of this approach.
%    Since the baseline models require four continuous frames, we input rendered frames $f_3, f_4, f_5$ and $f_6$ to predict $f_7$.
%    We recursively use the predicted frames to predict the subsequent frames.

    \autoref{fig:multistep-qualitative} compares example multi-frame predictions by DeCOMPnet with DPG and \autoref{fig:multistep-quantitative} shows average PSNR and SSIM for different models.
    We observe that DeCOMPnet outperforms all the competing models in terms of SSIM.
    In terms of PSNR, we are competitive with DPG in the prediction of $\hat{f}_{n+1}$ on our dataset, despite DPG additionally using $f_{n-1}, f_{n-2}$ and $f_{n-3}$.
    Further, DPG predictions are often blurry, which is not captured well by PSNR.
    % In terms of SSIM, which captures blur in the predicted frames, DeCOMPnet significantly outperforms DPG.
    Video comparisons are available in supplementary.

    \subsection{Ablations}\label{subsec:ablations}
    \textbf{2D vs 3D Flow Prediction:}
    We compare our 3D flow prediction model against a 2D flow model by predicting 2D flow between the frames $f_n$ and $f^w_{n-2}$.
    We use a model similar to the one described in \autoref{subsec:local-object-motion-prediction} on frames with 2D convolutions and cost volumes.
    Note that this model still uses partial convolutions and masked correlation layers to handle holes in $f^w_{n-2}$.
    We also feed depth as input to the flow estimation network for a fair comparison.
    Owing to different ranges of depth across multiple scenes, we first normalize depth to the range $[0,1]$, and then feed it to the flow estimation network.
    While the 2D model uses depth naively by concatenating depth with the input in an additional channel, the 3D model uses a more structured MPI representation.
    This comparison allows us to analyse the importance of using MPIs for flow estimation.
%    During training, along with the photometric loss on the reconstructed frame, we also impose an MAE loss on the reconstructed depth map, similar to \autoref{eq:loss-optical-flow-photometric}.

    We evaluate the flows predicted by 2D and 3D models using average endpoint error (AEPE)~\cite{liu2020arflow} for single frame prediction.
    For the test scenes in our dataset, we additionally render the optical flow corresponding to object motion alone and use it to compute the endpoint errors.
    Even though our model predicts 3D flow, we use the x-y components only to compute the AEPE\@.
    As argued earlier, estimating the object motion in 3D allows better matching of points, leading to a more accurate estimation of flow, even in x-y dimensions (see supplementary for more details).

    From \autoref{tab:ablations}, we observe that estimating the flow in 3D using MPI reduces AEPE by 38\%.
    Further, we observe in \autoref{fig:flow-2d-vs-3d} that the flow predicted by our 3D model is sharper leading to undistorted reconstructions at the edges, in contrast to 2D flow.

    \textbf{Impact of partial convolutions and masked correlations:}
    We study the impact of the partial convolution and masked correlation layers in DeCOMPnet by replacing them with standard 3D convolution and correlation layers.
    We evaluate the performance of object motion prediction using AEPE in \autoref{tab:ablations}.
    We observe that the proposed masked correlations and the use of partial convolutions to handle the sparsity in MPI representation lead to a significant improvement in the performance of object motion prediction.

    \subsection{Analysis of Performance Bounds}\label{subsec:analysis}
    \begin{table}
        \centering
        \caption{Performance bound analysis of different components of our model for single frame prediction.
        LMP: Local motion prediction; DI: Disocclusion infilling.
        Pred indicates flow prediction or infilling done by the network.
        GT indicates ground truth flow or infilling.
        }
        \begin{tabular}{cc|cc|cc}
            \hline
            \multirow{2}{*}{\textbf{LMP}} & \multirow{2}{*}{\textbf{DI}} & \multicolumn{2}{c|}{Our Dataset} & \multicolumn{2}{c}{MPI Sintel} \\
            & & \textbf{PSNR \textuparrow} & \textbf{SSIM \textuparrow} & \textbf{PSNR \textuparrow} & \textbf{SSIM \textuparrow} \\
            \hline
            pred & pred & 30.60 & 0.9314 & 29.64 & 0.8975 \\  % VPR016 Test0001, Test0101
            GT & pred & 30.67 & 0.9354 & 31.90 & 0.9426 \\  % VPR016 Test0002, Test0102
            pred & GT & 32.00 & 0.9377 & 30.35 & 0.9097 \\  % VPR016 Test0003, Test0103
            GT & GT & 33.53 & 0.9453 & 34.02 & 0.9613 \\  % VPR016 Test0004, Test0104
            \hline
        \end{tabular}
        \label{tab:bound-analysis}
    \end{table}

%    \begin{figure}
%        \centering
%        \includegraphics[width=\linewidth]{res/images/QualitativeComparisons_Ablations02}
%        \caption{Comparison of our disocclusion infilling with VINet based infilling.
%        The disoccluded region in the motion warped frame is shown in black.}
%        \label{fig:qualitative-ablations02}
%    \end{figure}

%    \textbf{Performance bounds:}
    We now analyze the upper bound on the performance of our model components for single frame prediction.
    We establish an upper bound on the performance that can be achieved by improving the object motion prediction by replacing the predicted total motion with the true optical flow provided by the graphics renderer.
    We warp $f_n$ with the ground truth optical flow to get $\hat{f}^w_{n+1}$ and then create $\hat{m}^w_{n+1}$ as explained in \autoref{subsec:mpi}, which is then fed to the disocclusion infilling module.
    To upper bound the performance that can be achieved by improving the disocclusion infilling, we apply alpha compositing on $\hat{m}^w_{n+1}$ and replace the disoccluded regions with true intensities from $f_{n+1}$.
    We also obtain a joint bound using both true optical flow and infilling with true intensities.
%    This gives an upper bound on the performance improvement that can be achieved by improving both object motion prediction and disocclusion infilling.

    We see from \autoref{tab:bound-analysis} that the performance of our model is close to the upper bound on our dataset.
    The larger gap in the MPI-Sintel database could be attributed to the challenging motion trajectories.
    The non-perfect reconstruction performance of the bound in the last row of \autoref{tab:bound-analysis} may be due to splatting approximations in warping.

    \subsection{Timing Analysis}\label{subsec:inference-time}
    Our model takes 4.5s to predict a single full HD frame on an Intel Core i7-9700F CPU with 32GB RAM and NVIDIA RTX 2080 Ti GPU, whereas Blender typically takes about 5m-1h to render a single frame depending on the scene.
    On further analysis, we find that the convolutional layers in optical flow estimation and disocclusion infilling take about 70ms and 3ms, respectively.
    Thus a significant amount of time in our implementation is consumed by warping operations.
    However, it is possible to optimize warping as shown by Barnes \etal~\cite{barnes2017positional} and Waveren \etal~\cite{van2016asynchronous}, which use less than 10ms.
    Further, due to the sparsity of the MPI representation, at any given location, alpha will be 0 on $Z-1$ planes.
    Although we ignore convolution layer outputs at locations where $\alpha=0$, inference time can be further reduced by $Z$ times by not convolving such points.
    With the above optimizations, the inference time of our model could reduce to less than 33ms, making it feasible for real-time use.

%     \subsection{Limitations}\label{subsec:failure-cases}
%     \textcolor{teal}{The superior performance of our model stems from its ability to separate the moving foreground objects from static background regions into different planes in the MPI representation.
%     However, when a moving object is closer in depth to a static background object, both the objects might end up on the same plane, which could lead to a drop in performance.
%     This is typically observed near the boundary between objects and the ground.
%     This problem could be alleviated by employing object segmentation along with depth to place objects onto different planes.
%     At lower frame rates, the linear motion model may be inefficient, and a deep prediction network might be useful.
% %    Further, when depth is unavailable, MPI may need to be predicted from the frames using a network~\cite{tucker2020single, li2021mine}.
% %    In future, we also plan to explore updating the MPI representation with the arrival of periodically rendered frames.
%     Lastly, since we do not predict new regions entering the scene, the frames have to be rendered at a larger resolution.
%     }

    \section{Conclusion}\label{sec:conclusion}
    In this work, we propose a novel framework for temporal view synthesis of dynamic scenes in the context of causal frame-rate upsampling of videos.
    We account for camera and object motion sequentially, which allows our framework to exploit the availability of camera motion effectively.
    Further, we estimate and predict object motion in the 3D MPI representation using masked correlations and partial convolutions.
    Finally, we infill disocclusions in the warped MPIs and use alpha-compositing to render the predicted frames.
    To evaluate our model, we develop a new dataset that brings out the challenges in temporal view synthesis.
    In future, we plan to extend our framework to real-world videos where ground truth depth may be unavailable, which would be useful in remote presence applications.

 \acknowledgments{
 This work was supported in part by a grant from Qualcomm. The first author was supported by the Prime Minister's Research Fellowship awarded by the Ministry of Education, Government of India.
 }

\appendix
\twocolumn[\subsection*{\centering \fontsize{13}{15}\selectfont Supplement}]
    \noindent The contents of this supplement include
    \begin{enumerate}[label=\Alph*., noitemsep]
        \item Video examples
        \item Use of MPI representation in 3D flow estimation
        \item Architecture details of optical flow estimation and disocclusion infilling
        \item Miscellaneous items
    \end{enumerate}

    \section{Video Examples}\label{sec:video-examples}
    While we compare the spatial quality of predictions by different models in Figs.\ 4 and 5 (in the main paper), we attach videos in this supplement to compare the temporal quality of predictions.
    Apart from our dataset, we additionally render a $5$s video of a scene from our dataset at $30$fps to qualitatively compare the performances of TVS models.

    \textbf{Single Frame Prediction:}
    Here alternate frames in the video are rendered and the intermediate frames are predicted using previous frames.
    To predict $f_{n+1}$, our model uses $f_n$ and $f_{n-2}$ whereas the benchmarked models use $f_n, f_{n-1}, f_{n-2}$ and $f_{n-3}$.
    Since rendered frames are available only for alternate time-instants, the predicted frames are recursively used to predict the subsequent frames \ie to predict $\hat{f}_{13}$, inputs to the benchmark video prediction models are $f_{12}, \hat{f}_{11}, f_{10}$ and $\hat{f}_9$.
    \begin{itemize}
        \item \path{city02_seq00_singleframe.mp4}: Comparison between predictions of DeCOMPnet and DPG~\cite{gao2019disentangling}, which is the second best performing model on our dataset.
        We also show the ground truth video for reference.
        Since the alternate frames are graphically rendered, distortions in the DPG predicted frames appear as flickering artifacts in the video.
        Whereas, the video predicted by our model is smoother with minimal artifacts.
        The video is played at 30fps.
    \end{itemize}

    \textbf{Multi Frame Prediction:}
    Here, we set $k=5$ to predict $\hat{f}_{n+1}, \hat{f}_{n+2}, \hat{f}_{n+3}$ and $\hat{f}_{n+4}$ given $f_n$ and $f_{n-5}$.
    Thus, in all the below videos, every 5th frame is rendered and all the intermediate frames are predicted.
    \begin{itemize}
        \item \path{city02_seq00_singleframe.mp4}: Compares the predictions of DeCOMPnet and DPG\@.
        Notice that DPG fails to predict proper motion, while our model is able to predict it reasonably well.
        The video is played at 30fps.
        \item \path{shaman3_albedo_multiframe.mp4}: Compares the predictions of DeCOMPnet and PreCNet.
        To clearly observe both object and camera motion between the frames, this video is played at 1fps.
        The video contains only five frames where the first frame is rendered and the next four frames are predicted.
    \end{itemize}

    % We request the reader to use the attached html file which provides options to play the videos.
    % We observed that web browsers, like chrome or safari, are able to play low frame-rate videos (1fps) better than default video players.
    % So, we request the reader to play such videos in a web browser.
    % However, the videos at 30fps can be played by any video player or web browser.
    Videos are available on our project webpage \url{https://nagabhushansn95.github.io/publications/2022/DeCOMPnet.html}.

    \section{Use of MPI representation in 3D flow estimation}\label{sec:mpi-flow-estimation}
    We observe that MPI representations are more useful than RGB-D representations in determining the flow in the occluded regions.
    Determining the flow accurately in these regions is vital for predicting the future frames with minimal distortions.
    For example, consider a region in $f_n$ that is occluded in $f^w_{n-k}$.
    Recall that occluded regions usually belong to the relative background.
    In such occluded regions, there are no matching pixels in the pair of frames, and luminance constancy fails.
    Thus, flow estimation in such regions is usually guided by the flow in its neighborhood through smoothness constraints.
    The neighborhood of an occluded region in an RGB-D representation contains both the relative foreground and background.
    Hence, the flow estimated in such regions is a combination of the foreground and background flow, leading to distortions in the predicted frame (\autoref{fig:flow-2d-vs-3d}).
    However, the neighborhood of such occluded regions in an MPI plane contains only the relative background;
    hence, the estimated flow depends only on the flow of the background.

    We further visualize this benefit in \autoref{fig:flow-2d-vs-3d}.
    In the example in the first row, the car (foreground) moves to the left, whereas the building (background) is stationary.
    The portion of the building to the immediate right of the car is visible in frame n, but hidden in frame n-2.
    The figure shows that the flow estimated in this occluded region with RGB-D representation is non-zero.
    As a result, this region of the building gets distorted (blue arrow) in the predicted next frame.
    However, the flow estimated with MPI representation is zero in this region, leading to an undistorted reconstruction of this region in the motion predicted frame.

    \section{Architecture details of Optical Flow Estimation and Disocclusion Infilling}\label{sec:flow-math}
    \begin{table*}
        \centering
        \begin{tabular}{l|lccccc}
            \hline
            \textbf{id} & \textbf{layer} & \textbf{kernel size} & \textbf{no. of output filters} & \textbf{stride} & \textbf{padding} & \textbf{activation} \\
            \hline
            1a & p-conv3d & (3, 3, 3) & 16 & (2, 2, 1) & (1, 1, 1) & Leaky ReLU \\
            1b & p-conv3d & (3, 3, 3) & 16 & (1, 1, 1) & (1, 1, 1) & Leaky ReLU \\
            2a & p-conv3d & (3, 3, 3) & 32 & (2, 2, 1) & (1, 1, 1) & Leaky ReLU \\
            2b & p-conv3d & (3, 3, 3) & 32 & (1, 1, 1) & (1, 1, 1) & Leaky ReLU \\
            3a & p-conv3d & (3, 3, 3) & 64 & (2, 2, 1) & (1, 1, 1) & Leaky ReLU \\
            3b & p-conv3d & (3, 3, 3) & 64 & (1, 1, 1) & (1, 1, 1) & Leaky ReLU \\
            4a & p-conv3d & (3, 3, 3) & 96 & (2, 2, 1) & (1, 1, 1) & Leaky ReLU \\
            4b & p-conv3d & (3, 3, 3) & 96 & (1, 1, 1) & (1, 1, 1) & Leaky ReLU \\
            5a & p-conv3d & (3, 3, 3) & 128 & (2, 2, 1) & (1, 1, 1) & Leaky ReLU \\
            5b & p-conv3d & (3, 3, 3) & 128 & (1, 1, 1) & (1, 1, 1) & Leaky ReLU \\
            6a & p-conv3d & (3, 3, 3) & 192 & (2, 2, 1) & (1, 1, 1) & Leaky ReLU \\
            6b & p-conv3d & (3, 3, 3) & 192 & (1, 1, 1) & (1, 1, 1) & Leaky ReLU \\
            \hline
        \end{tabular}
        \caption{Details of the feature extraction network for optical flow estimation.}
        \label{tab:pwc-feature-extraction}
    \end{table*}

    \begin{table*}
        \centering
        \begin{tabular}{l|lcccccc}
            \hline
            \textbf{id} & \textbf{layer} & \textbf{kernel size} & \textbf{no. of output filters} & \textbf{skip connection} & \textbf{stride} & \textbf{padding} & \textbf{activation} \\
            \hline
            1 & p-conv3d & (3, 3, 3) & 128 & - & (1, 1, 1) & (1, 1, 1) & Leaky ReLU \\
            2 & p-conv3d & (3, 3, 3) & 128 & - & (1, 1, 1) & (1, 1, 1) & Leaky ReLU \\
            3 & p-conv3d & (3, 3, 3) & 96 & 1 & (1, 1, 1) & (1, 1, 1) & Leaky ReLU \\
            4 & p-conv3d & (3, 3, 3) & 64 & 2 & (1, 1, 1) & (1, 1, 1) & Leaky ReLU \\
            5 & p-conv3d & (3, 3, 3) & 32 & 3 & (1, 1, 1) & (1, 1, 1) & Leaky ReLU \\
            6 & p-conv3d & (3, 3, 3) & 2 + (2 $s_z$ + 1) & 4 & (1, 1, 1) & (1, 1, 1) & Linear(2) + Softmax($2 s_z + 1$) \\
            \hline
        \end{tabular}
        \caption{Details of the the network that estimates flow from cost volume. Skip connection: id of the layer whose output is concatenated with the input at the current layer.}
        \label{tab:pwc-flow-estimation}
    \end{table*}

    \begin{table*}
        \centering
        \begin{tabular}{l|lcccccc}
            \hline
            \textbf{id} & \textbf{layer} & \textbf{kernel size} & \textbf{no. of output filters} & \textbf{skip connection} & \textbf{stride} & \textbf{padding} & \textbf{activation} \\
            \hline
            1 & conv3d & (7, 7, 7) & 32 & - & (1, 1, 1) & (3, 3, 3) & ReLU \\
            2 & conv3d & (5, 5, 5) & 64 & - & (2, 2, 1) & (2, 2, 2) & ReLU \\
            3 & conv3d & (3, 3, 3) & 128 & - & (2, 2, 1) & (1, 1, 1) & ReLU \\
            4 & conv3d & (3, 3, 3) & 128 & - & (2, 2, 1) & (1, 1, 1) & ReLU \\
            5 & conv3d & (3, 3, 3) & 128 & - & (2, 2, 1) & (1, 1, 1) & ReLU \\
            6 & conv3d & (3, 3, 3) & 128 & 4& (0.5, 0.5, 1) & (1, 1, 1) & ReLU \\
            7 & conv3d & (3, 3, 3) & 64 & 3& (0.5, 0.5, 1) & (1, 1, 1) & ReLU \\
            8 & conv3d & (3, 3, 3) & 32 & 2& (0.5, 0.5, 1) & (1, 1, 1) & ReLU \\
            9 & conv3d & (3, 3, 3) & 2 & 1& (0.5, 0.5, 1) & (1, 1, 1) & Linear \\
            \hline
        \end{tabular}
        \caption{Details of the disocclusion infilling network. Skip connection: id of the layer whose output is concatenated with the input at the current layer. Fractional strides represent upsampling followed by skip connection and the convolution layer.}
        \label{tab:network-architecture-infilling}
    \end{table*}

    We use PWC-Net backbone for optical flow estimation.
    \begin{enumerate}
        \item We first extract 3D features from the MPIs $m_n$ and $m^w_{n-2}$ using the network described in \autoref{tab:pwc-feature-extraction}.
        \item The estimated flow from previous scale is upsampled by 2.
        At the lowest scale, previous flow is taken as zero.
        \item The features of $m^w_{n-2}$ at the given scale are warped using the upsampled flow.
        \item The features of $m_n$ and the warped features of $m^w_{n-2}$ at the given scale are passed through a masked correlation layer (Eq.3 and Eq.4) to compute a cost volume.
        \item The features of $m_n$ are also processed with a $1 \times 1 \times 1$ partial convolution layer to reduce the number of channels to 32.
        \item The cost volume from step 3 and the processed features from step 4 are concatenated along with the flow estimated from previous scale.
        The concatenated volume is then fed to the network described in \autoref{tab:pwc-flow-estimation} to estimate the residual flow at the given scale.
        Output activation is linear for the first two channels and softmax on the last ($2 s_z + 1$) channels.
        Note that these flow estimation layers share the weights across all the scales.
        \item Steps 2--6 are repeated for each scale except the last two scales.
        In the last two scales, the flow from previous scale is simply upsampled without computing any residual flow.
    \end{enumerate}

    As is typical in multiscale flow estimation, we estimate absolute flow in only the lowest scale and estimate only the residual flow in higher scales.
    Let $(\mathbf{a}_1, b_1)$ be the estimated flow from the previous scale and $(\tilde{\mathbf{a}}_2, \tilde{b_2})$ be the residual flow estimated in the subsequent scale.
%    We first expand the $s_z$ dimensional flow vector $\tilde{w}^z$ estimated at location $\left(\mathbf{x}, p\right)$ to $P$ dimensions by padding zeros corresponding to the planes outside the search field used in cost volume computation to get $\bar{w}_2 \in \mathbb{R}^{h \times w \times P \times (2 + P)}$.
    We compute the effective flow, $(\mathbf{a}_2, b_2)$, as
    \begin{align}
        \mathbf{a}_2 \left( \mathbf{x}, z \right) &= \mathbf{a}_1 \left( \mathbf{x}, z \right) + \tilde{\mathbf{a}}_2 \left( \mathbf{x}, z \right) \\
        b_2 \left( \mathbf{x}, z \right) &= \mathbb{E}_{\tilde{b}_2} \left[ b_1 \left( \mathbf{x} + \tilde{\mathbf{a}}_2\left( \mathbf{x}, z \right), z + z' \right) \right],
    \end{align}
    where $z'$ is the estimated difference in the index of the MPI planes as described in Sect.\ 4.3.
    $\mathbb{E}$ is the expectation operator.
    The flow at the final scale is then converted to a real-valued 3D flow using Eq.\ 5.

    \autoref{tab:network-architecture-infilling} shows the architecture of the U-Net used for disocclusion infilling.

    \section{Miscellaneous Items}\label{sec:misc}
    \subsection{Choice of Hyper Parameters}
    \begin{itemize}
        \item We set the number of MPI planes $Z = 4$ due to memory constraints in our implementation.
        A higher value of $Z$ would likely improve the performance.
        As argued in Sect.\ 5.8, with sparse convolutions, increasing $Z$ would have little impact on memory and speed.
        \item We empirically find that $s_z = 1$ gives reasonable performance.
        \item $\beta$ in Eq.\ 7 balances the MAE and SSIM loss and we set it to 0.15, as common in the literature~\cite{godard2017unsupervised}.
        \item $a$ in Eq.\ 10 controls the level of smoothness constraint on the estimated flow.
        We set $a=10$ as suggested by Liu \etal~\cite{liu2020arflow}.
        \item $\lambda$ in Eq.\ 11 balances the two loss terms.
        Through a coarse grid search, we set $\lambda = 10$.
        \item During inference, we run disocclusion infilling iteratively for $g=3$ following Kanchana \etal~\cite{kanchana2022ivp}.
    \end{itemize}

\bibliographystyle{abbrv-doi}

\bibliography{VPTU}
\end{document}